\documentclass[conference]{IEEEtran}
\IEEEoverridecommandlockouts
\usepackage{cite}
\usepackage{amsmath,amssymb,amsfonts}
\usepackage{algorithmic}
\usepackage{graphicx}
\usepackage{textcomp}
\usepackage{xcolor}
\def\BibTeX{{\rm B\kern-.05em{\sc i\kern-.025em b}\kern-.08em
    T\kern-.1667em\lower.7ex\hbox{E}\kern-.125emX}}
\usepackage{enumitem}
\usepackage{url}
\usepackage{listings}
\usepackage{xcolor}

\definecolor{codegreen}{rgb}{0,0.6,0}
\definecolor{codegray}{rgb}{0.5,0.5,0.5}
\definecolor{codepurple}{rgb}{0.58,0,0.82}
\definecolor{backcolour}{rgb}{0.95,0.95,0.92}

\lstdefinestyle{mystyle}{
    backgroundcolor=\color{backcolour},   
    commentstyle=\color{codegreen},
    keywordstyle=\color{magenta},
    numberstyle=\tiny\color{codegray},
    stringstyle=\color{codepurple},
    basicstyle=\ttfamily\footnotesize,
    breakatwhitespace=false,         
    breaklines=true,                 
    captionpos=b,                    
    keepspaces=true,                 
    numbers=left,                    
    numbersep=5pt,                  
    showspaces=false,                
    showstringspaces=false,
    showtabs=false,                  
    tabsize=2
}

\usepackage{algorithm2e}
\RestyleAlgo{ruled}
\usepackage{float}

\begin{document}

\title{LLM Enhancer: Merged Approach using Vector
Embedding for Reducing Large Language Model
Hallucinations with External Knowledge\\
%%{\footnotesize \textsuperscript{*}Note: Sub-titles are not captured in Xplore and should not be used}
%%\thanks{Identify applicable funding agency here. If none, delete this.}
}
\author{\IEEEauthorblockN{Naheed Rayhan${^1}$, Md. Ashrafuzzaman${^1}$}
\IEEEauthorblockA{${^1}$Department of Computer Science and Engineering, Jagannath University, Bangladesh\\
Email: \{naheed27ray, naheed28ray\}@gmail.com, mdashrafuzzamanabir@gmail.com}}

\maketitle

\begin{abstract}
Large Language Models (LLMs), such as ChatGPT, have demonstrated the capability to generate human-like, natural responses across a range of tasks, including task-oriented dialogue and question answering. However, their application in real-world, critical scenarios is often hindered by a propensity to produce inaccurate information and a limited ability to leverage external knowledge sources. Furthermore, the reliance on outdated training data can impede their accuracy.
This paper introduces the LLM-ENHANCER system, designed to integrate multiple online sources—such as Google, Wikipedia, and DuckDuckGo—to enhance data accuracy. The LLMs employed within this system are open-source. The data acquisition process for the LLM-ENHANCER system operates in parallel, utilizing custom Agent tools to manage the flow of information, as opposed to a serial approach. Vector embeddings are employed to discern the most pertinent information, which is subsequently supplied to the LLM for user interactions.The LLM-ENHANCER system notably mitigates the incidence of hallucinations in chat-based LLMs, while preserving the naturalness and accuracy of their responses.\\
\end{abstract}


\section{Introduction}

\subsection{Background}
 In the realm of natural language processing, the transformer architecture is a deep learning approach that leverages a multi-head attention mechanism. It was introduced in a paper by Ashish Vaswani, Noam Shazeer et al. in 2017 \cite{vaswani2023attention}.\\

The architecture underlying Large Language Models (LLMs), such as GPT-3 \cite{brown2020language} and ChatGPT, serves as the foundation for their ability to generate natural language texts that are fluent, coherent, and informative. These models have demonstrated exceptional performance, widely attributed to their extensive knowledge of the world, enabling effective generalization. Nevertheless, it is crucial to acknowledge that the encoding of knowledge in LLMs is not infallible, resulting in potential memory distortions and hallucinations \cite{alkaissi2023hallucinations}. This becomes particularly concerning when deploying these models for critical tasks. Thus, it is essential to take into account the limitations of LLMs and consider their potential drawbacks when employing them in sensitive contexts.\\

This paper proposes an LLM-ENHANCER system that merges multiple online
sources, using agents for extracting data from Google, Wikipedia, and DuckDuckGo, to obtain more
accurate data. The LLMs utilized in this system are open-source. The data is merged and then split into chunks to pass it to the vector embedding database. Vector embedding is used to identify the most
relevant chunk of information, which is provided to the LLM for answering the question of the user.

\subsection{Motivation}

Large Language Models (LLMs) have revolutionized the landscape of artificial intelligence, profoundly impacting various industries and societal domains. Their advanced natural language understanding capabilities have enabled applications ranging from automated customer service and personalized content generation to language translation and medical diagnostics. \\

The knowledge base of machine learning models (LLMs) is vast and complex, yet it is not immune to imperfections, and can sometimes generate inaccurate or distorted information. This issue is heightened by the models' susceptibility to producing false data, which poses a considerable risk in mission-critical scenarios. The concept of "memory distortion" refers to the potential inaccuracies that may arise during the encoding and generalization processes, resulting in deviations from accurate information. In essence, memory distortion is a critical concern for organizations that rely on machine learning models as it can lead to unintended consequences and significant losses. Therefore, it is essential to continuously evaluate the accuracy and reliability of LLMs to mitigate the risks associated with memory distortion.\\

The issue at hand is a significant obstacle that requires urgent attention. Finding a solution would not only resolve the problem but also improve the overall process. Furthermore, the expense of fine-tuning LLM for custom data serves as a motivation to seek alternative approaches that are more cost-effective.

\subsection{Problem Statement}

Machine learning models have a vast and intricate knowledge base, but they are not perfect and can sometimes provide incorrect or distorted information. This issue is worsened by the models' tendency to produce false data, which can be a significant risk in mission-critical situations. "Memory distortion" refers to the possibility of inaccuracies that may occur during encoding and generalization, resulting in deviations from accurate information. Organizations that rely on machine learning models must be aware of memory distortion as it can lead to unintended consequences and significant losses. Therefore, it is crucial to continuously assess the accuracy and dependability of machine learning models to mitigate the risks associated with memory distortion.\\

The present matter that we are dealing with poses a significant challenge that requires immediate attention. Finding a viable solution to this issue would not only help us overcome the obstacle but also optimize the entire process. This, in turn, would lead to improved outcomes, increased efficiency, and better customer satisfaction. Additionally, the cost of fine-tuning LLM for custom data is quite high, which makes it impractical for us to continue with this approach. Therefore, we must explore alternative approaches that are more cost-effective and efficient. By doing so, we can achieve our goals while also keeping the expenses under control.

\subsection{Contributions}

Baolin Peng, Michel Galley, et al. \cite{peng2023check} presented a paper where they proposed a solution to reduce the hallucinations of the LLM. They developed a system called LLM-AUGMENTER with Plug and Play (PnP) modules, one of which is "Policy". The "Policy" module selects the next system action \cite{peng2023check}.\\

After the developers shared the module's architecture, the online community built a similar system called Agent Executor, which takes action based on an agents tool description.\\

Agent Executor \cite{langchain_agents_concepts} takes action based on the description of the tools that are being provided. Now if there are multiple sources of homogeneous data the tool selection is random and depends solely on the provided description of the tools.\\

We are suggesting the use of an LLM-ENHANCER system that combines information from various online sources using agents to extract data from Google, Wikipedia, and DuckDuckGo. The LLM used in this system is open-source. To process the data, it is combined and broken into smaller parts, which are then passed through a vector embedding. The vector embedding identifies the most relevant information and provides it to the LLM to answer the user's question. This technique is used to remove the hallucination of the LLMs.\\

Our LLM-ENHANCER empowers LLM models to provide accurate information without extensive training. With this model, we require less computational power since we do not need to train it with big data to obtain accurate information. \\
%=========================================================

\subsection{Organization}
The remainder of this paper is organized as follows:\\
\begin{itemize}
  \item \textbf{Chapter 1 Introduction:} In this chapter, we introduce our thesis problem, the motivation behind selecting this field and our contributions to the field.
  \item \textbf{Chapter 2  State-of-the-Art-Works:} In this chapter, we discussed the prior
works related to this field. We provided information on previous techniques that were used by others. 
  \item \textbf{Chapter 3 Proposed Methodology:} In this chapter, we described our main proposed model for LLM-ENHANCER.
  \item \textbf{Chapter 4 System Implementation:} In this chapter,  we present the system
implementation, briefly describing the techniques used in this study and what tools we have used.
  \item \textbf{Chapter 5 Performance Evaluation:} In this chapter, we discuss the evaluation metrics and comparisons of the models' results.
  \item \textbf{Chapter 6 Conclusions and Future Work:} This chapter presents definitive conclusions and forward-thinking ideas for future work. Our proposed model also has certain limitations that must be considered in this chapter.
  
\end{itemize}

\vspace{1cm}

% ====================Chapter 2 ==========================================
\section{State-of-the-Art-Works}
In the field of natural language processing, various approaches have been proposed to enhance the performance of Large Language Models (LLMs).\\ 

Language models like ChatGPT are proficient in generating human-like and coherent responses across various tasks, including dialogues focused on specific tasks and answering questions. However, it becomes challenging to utilize LLMs in critical, real-world applications since they tend to generate inaccurate information, also known as hallucinations \cite{alkaissi2023hallucinations}, and face difficulties integrating external knowledge.\\

So, previous researchers explored various techniques and approaches based on custom tools, prompt engineering, database management, and actions to improve LLMs' knowledge base and external knowledge base.\\  

\subsection{LLM-AUGMENTER Approach}

Baolin Peng, Michel Galley et al. \cite{peng2023check} discussed LLM-AUGMENTER\cite{peng2023check} which improves a fixed LLM by consolidating evidence from external knowledge for the LLM to generate responses grounded in evidence and revising LLM's (candidate) responses using automated feedback. \\

It introduces LLM-AUGMENTER, a method to enhance fixed LLMs using external knowledge and automated feedback without parameter fine-tuning. LLM-AUGMENTER retrieves evidence from external sources, consolidates it, and queries the fixed LLM to generate responses grounded in evidence. It then verifies and revises the generated responses iteratively to reduce hallucinations. The paper aims to validate the effectiveness of LLM-AUGMENTER through empirical studies on information-seeking dialog and open-domain Wiki question-answering tasks, demonstrating significant reductions in hallucinations while maintaining fluency and informativeness of the generated responses. \\    

Baolin Peng, Michel Galley, et al. \cite{peng2023check} proposed mainly this LLM-AUGMENTER, a framework designed to enhance closed-source LLMs by incorporating external knowledge and automated feedback. The framework utilizes external knowledge in prompts to ground responses in relevant information and employs automated feedback to improve response quality.\\

Baolin Peng, Michel Galley, et al. \cite{peng2023check} have identified a major limitation in using models like ChatGPT, which is the potential decrease in user experience due to the high computational cost of interacting with the model. However, they suggest allowing users to choose between a faster initial response and a more accurate one to mitigate this issue. Another limitation is that the policy used in the experiments was manually designed, as training with reinforcement learning was deemed inefficient due to the high demand for ChatGPT. While the authors plan to conduct RL experiments with ChatGPT in the future, the current results rely on a policy trained with T5-Base. Additionally, the authors note that the current version lacks human evaluation and plan to include it in future iterations, especially as they explore more nuanced utility functions like safety.\\

\begin{figure}[h]
  \centering
  \includegraphics[width=0.75\columnwidth]{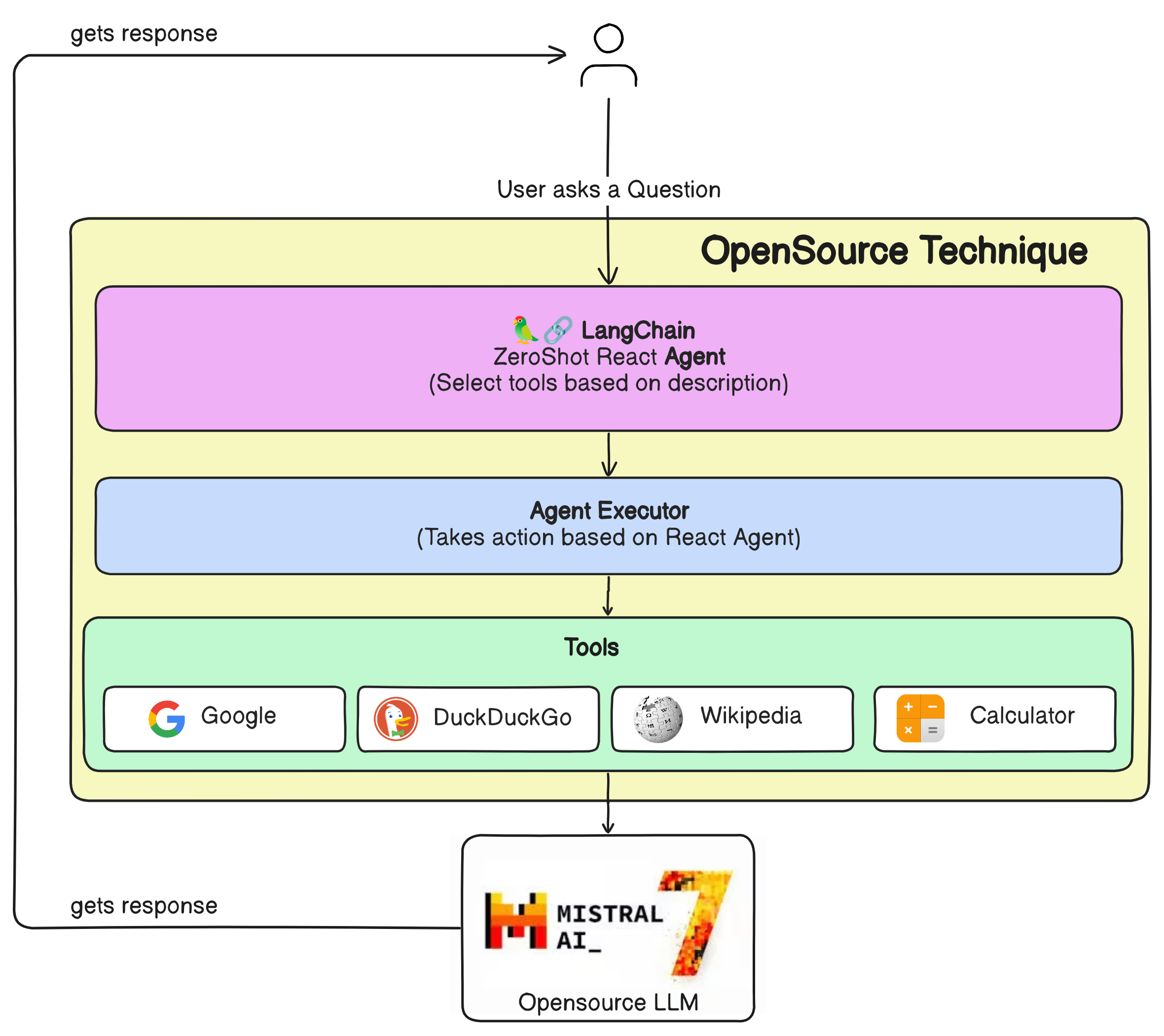}
  \caption{Widely used Open-Source Technique.}
  \label{fig:Widely used Open-Source technique}
\end{figure}

Figure \ref{fig:Widely used Open-Source technique} illustrates the open-source method of connecting LLM to external knowledge.

This figure \ref{fig:Widely used Open-Source technique} is a widely used Open-Source technique for retrieving homogeneous data using agents from the web. When a user asks questions to the LLM, it triggers the ZeroShot React Agent, which in turn triggers the Agent Executor. The Agent Executor randomly selects data from one tool and then sends all the information by a prompt to the LLM to get an answer. This is also known as prompt engineering. Here tools can be search engines, websites, databases,calculators, etc.

\subsection{Limitation of the existing works}

Baolin Peng, Michel Galley et al. \cite{peng2023check} and their colleagues have published a paper discussing LLM-AUGMENTER, which enhances Large Language Models using external knowledge and automated feedback. While they have provided a detailed explanation of the process, they have not released any code. Moreover, the policy module being used utilizes RL for selecting the next policy, which can be easily achieved using agents. Furthermore, they rely on a closed-source model. \\

However, a recent open-source project has suggested that specific tools can be used in conjunction with agents in a sequential manner. Our paper addresses this issue by using these tools in parallel, combining the data from multiple sources and forwarding it to a vector embedding database, such as ChromaDB \cite{chroma}. We then retrieve the required data and pass it on to the LLMs. By merging the tools through vector embeddings, we can achieve significant improvements.\\

Our approach offers a promising solution to the issue of sequential tool usage where the tools are selected based on Tools description. If the tools are homogenous then selecting the wrong tools may lead to the wrong result. By using vector embeddings and ChromaDB, we can merge most of the tools available for web interfacing and enhance the performance of LLMs.

\vspace{1cm}
\section{Proposed Methodology}

This section outlines our proposed model, LLM-ENHANCER, which enhances open-source LLMs and reduces hallucinations as shown in Figure \ref{fig:LLM-ENHANCER}. The section also provides an overview of the various modules that were used to develop the LLM-ENHANCER.

\begin{enumerate}[label=\textbf{\Alph*.}]
        \item LLM-ENHANCER 
        \begin{enumerate}[label=\textbf{\arabic*.}]
                \item ZeroShot React Agent 
                \item Action Executor 
                \item LangChain Tools 
                \item Merged Tools 
                \item Calculator 
                \item Splitter 
                \item Embeddings 
                \item ChromaDB 
                \item Mistral 7B (Opensource LLM) 
        \end{enumerate}
\end{enumerate}

\subsection{LLM-ENHANCER}
\begin{figure}[h]
  \centering
  \includegraphics[width=0.75\columnwidth]{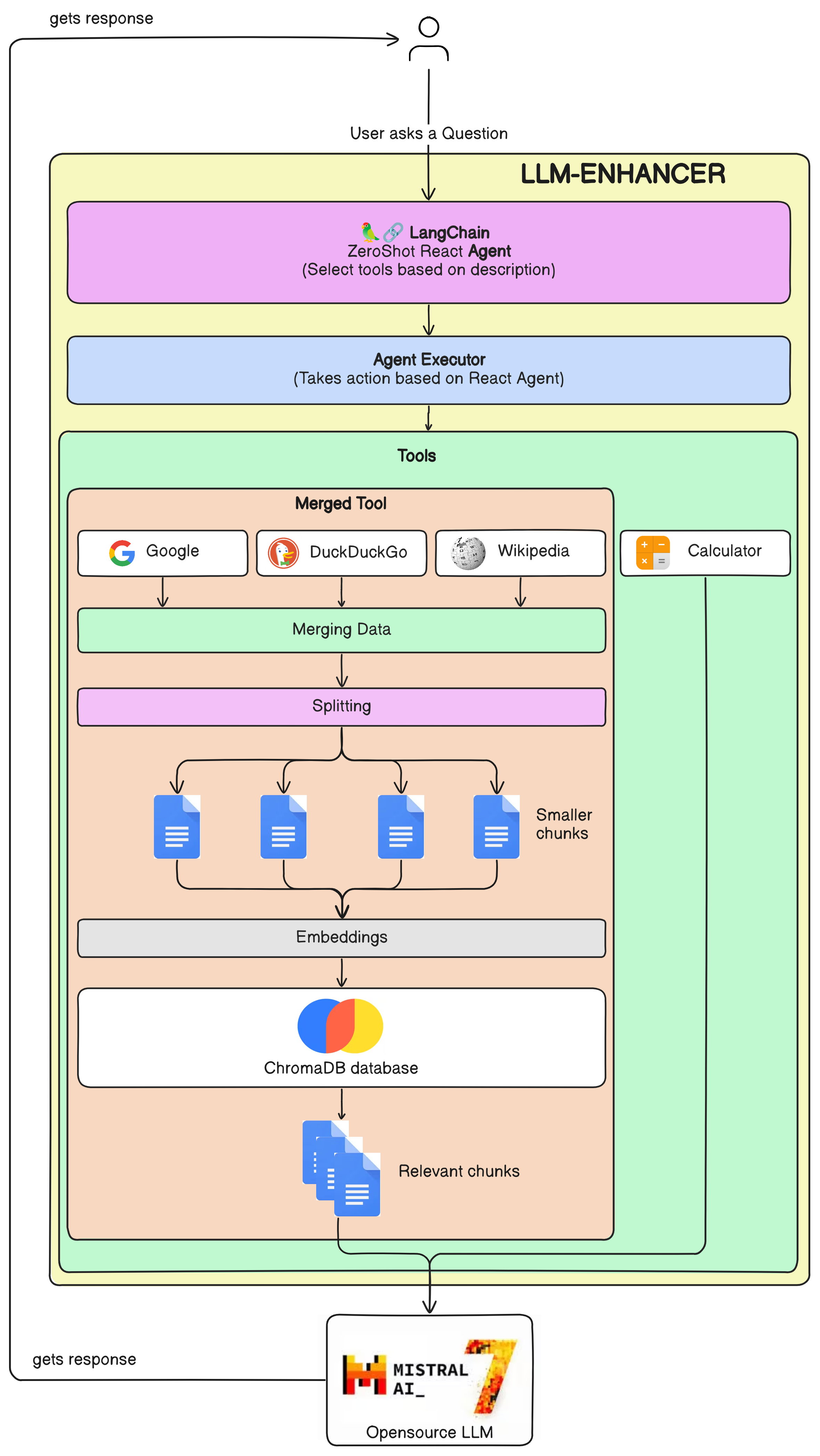}
  \caption{LLM-ENHANCER.}
  \label{fig:LLM-ENHANCER}
\end{figure}

The LLM-ENHANCER as shown in Figure \ref{fig:LLM-ENHANCER}, is a tool that takes input from the user, and the Agent Executor selects appropriate tools based on the description provided. However, all the homogenous tools are merged into a named Merged tool, which allows them to work as one. The data from multiple sources is combined in the Merged tool and then split into smaller chunks. These chunks are then passed to vector embeddings, which store the vector in the chromadb database. Since the chunks are stored in vector embeddings, we can obtain relevant chunks that are passed to the LLM for processing. This approach allows the LLM to produce much better results.

This figure \ref{fig:LLM-ENHANCER} is our proposed technique for extracting data from the web by utilizing Vector embedding for data accuracy.

In recent times, there has been a surge in the development of open-source models such as Llama \cite{touvron2023llama}, Llama2 \cite{touvron2023llama2}, Mistral7b \cite{jiang2023mistral}, and Bart \cite{lewis2019bart}. These models have brought about an entire ecosystem, including LangChain \cite{langchain} (a framework created to simplify the process of building applications with large language models) and GPT4All \cite{gpt4all} (an ecosystem that enables the deployment of powerful and customized large language models locally on consumer-grade CPUs and GPUs).\\

Nomic AI \cite{nomicai} is a leading provider of open-source models that are used in the GPT4ALL software ecosystem. These models are built on top of "mistralai/Mistral-7B-v0.1" \cite{mistralai} and range in size from 3GB to 8GB. Users can download and integrate these models into the GPT4ALL open-source ecosystem software. For our purposes, we have utilized the "mistral-7b-openorca.Q4\_0.gguf" \cite{mistral7bopenorca} open-source model. To further enhance its capabilities, GPT4ALL has been integrated with Langchain, which utilizes Agents \cite{langchain_agents} to connect to the internet and the web.\\

To enhance the performance of Large Language Models (LLMs), we have implemented LLM-ENHANCER. This approach involves an agent containing multiple tools to execute various tasks.
\begin{itemize}
    \item SerpAPI \cite{serpapi_integration} for Google search
    \item DuckDuckGo Search \cite{ddg_integration}
    \item Wikipedia \cite{wikipedia_integration}
\end{itemize}
The mentioned tools are not used independently. Instead, the data is combined and forwarded to a vector embedding database such as ChromaDB \cite{chroma}. After that, the required data is retrieved and passed on to the LLMs. By merging the tools through vector embeddings, we can make significant improvements.\\

\vskip1ex
\begin{figure}[h]
  \centering
  \includegraphics[width=0.8\columnwidth]{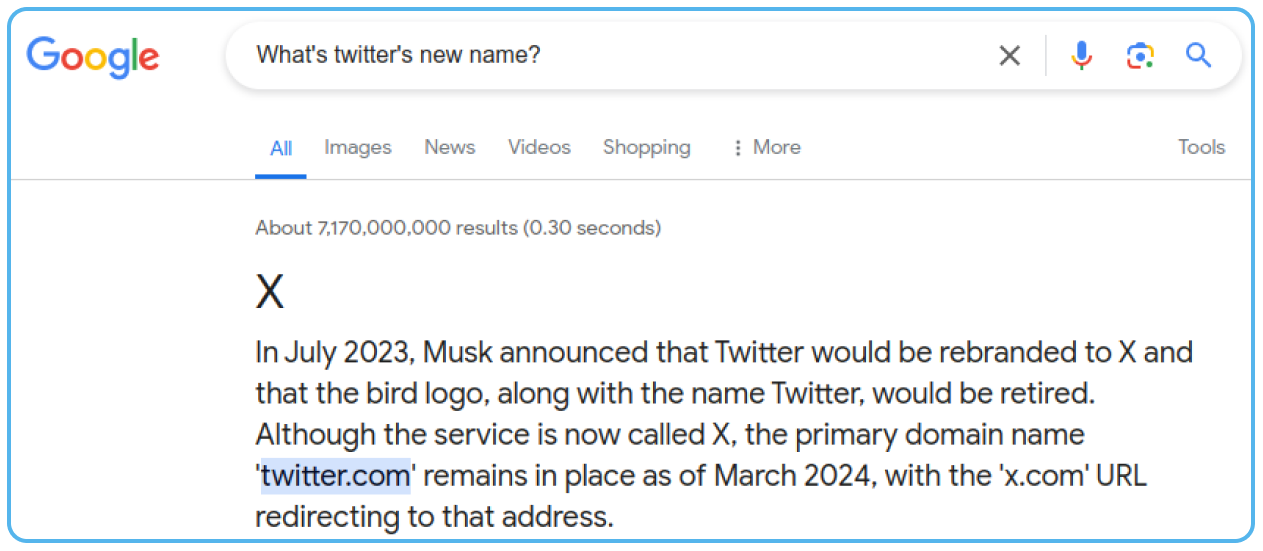}
  \caption{Real Answer from the web.}
  \label{fig:real1}
\end{figure}
\vskip2ex
\vskip1ex
\begin{figure}[h]
  \centering
  \includegraphics[width=0.8\columnwidth]{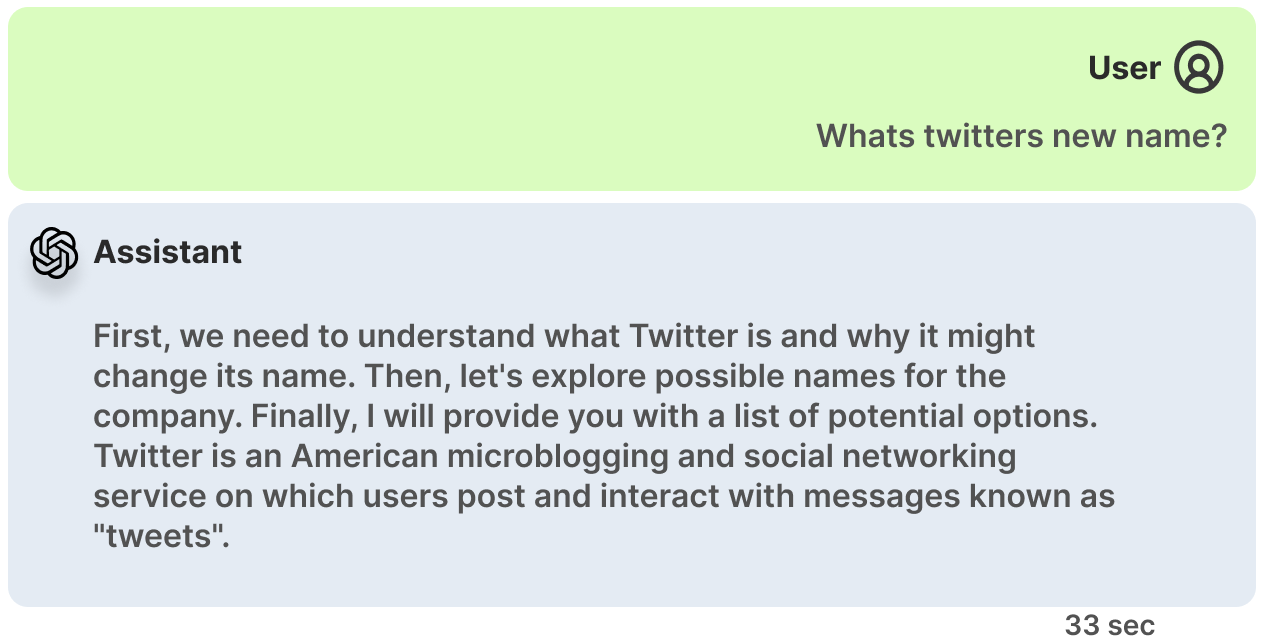}
  \caption{Answer from the LLM without internet access}
  \label{fig:internet1}
\end{figure}
\vskip2ex
\vskip1ex
\begin{figure}[h]
  \centering
  \includegraphics[width=0.8\columnwidth]{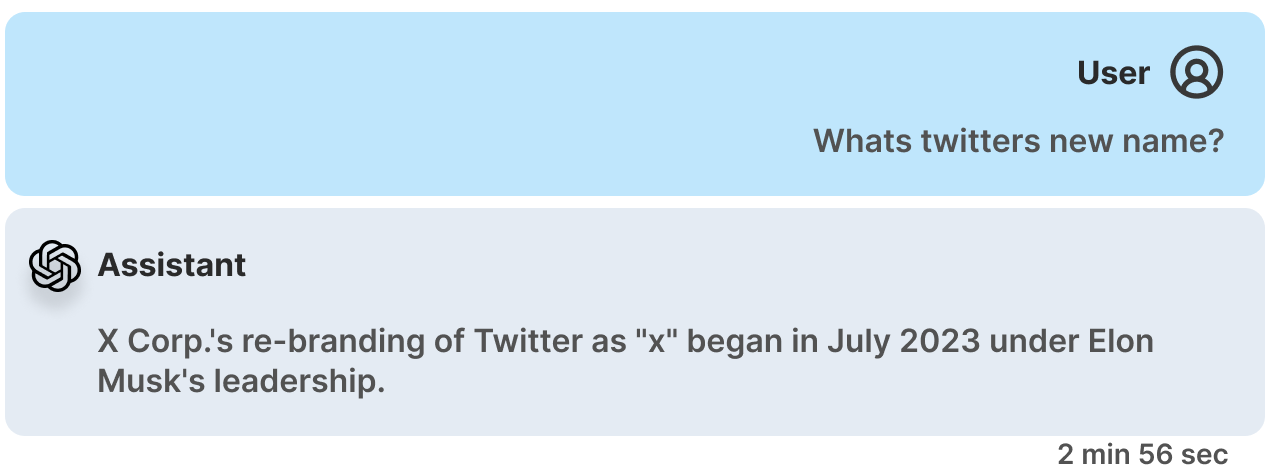}
  \caption{Answer from the agents where tools are Sequential}
  \label{fig:seq1}
\end{figure}
\vskip2ex
\vskip1ex
\begin{figure}[h]
  \centering
  \includegraphics[width=0.8\columnwidth]{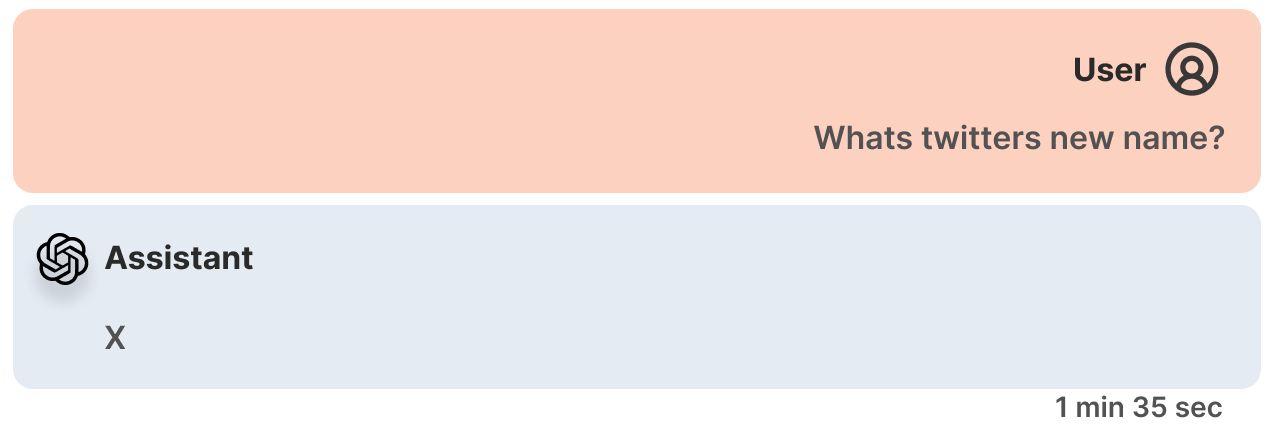}
  \caption{Answer from the agent where multiple tools are merged}
  \label{fig:merge1}
\end{figure}
\vskip2ex

It is evident from the provided information that the LLM's training data only extends up to October of 2021. Therefore, it would not have been privy to any updates on the social media platform's name change in 2023, as depicted in Figure \ref{fig:real1}. It is unsurprising, then, that the LLM was unable to provide a correct response to the question posed in Figure \ref{fig:internet1}.\\

In Figure \ref{fig:seq1}, by sequentially incorporating three tools-SerpApi (Google search) \cite{serpapi_integration}, DuckDuckGo search \cite{ddg_integration},wikipidea \cite{wikipedia_integration} and a calculator we successfully obtained accurate information within 2 minutes and 56 seconds.Here we have used laptop grade CPU(Intel Core i5-8250U)\\

To obtain data quickly and efficiently, we have used ChromaDB for vector embedding. Using Google search, DuckDuckGo search, and Wikipedia tools, we have collected the required information and combined the output of these tools into a single chunk passed to the chromadb \cite{chroma}. This allowed us to use vector embeddings to extract the most relevant information based on the user's question. The LLM was then used to process this data and find the necessary information.In Figure \ref{fig:merge1}, We have used a parallel merging technique, which improves efficiency compared to using each tool sequentially. As a result, we obtained the required information in just 1 minutes and 35 seconds, which is faster then Sequential Method.\\

LLMs are commonly used, however, they have some limitations. They cannot encode all the necessary information required for various applications, and they can become outdated. These limitations are further compounded by the constantly changing real-world settings and the lack of certain datasets for training. In these cases, this LLM-ENHANCER could be used.\\

Different efforts have been made to integrate external knowledge into LLMs, but many methods require expensive fine-tuning of model parameters. This is especially true as the sizes of LLMs increase. The LLM-ENHANCER approach introduces plug-and-play (PnP) modules that can be used to enhance a pre-existing LLM for mission-critical tasks without extensive parameter adjustments. This approach aims to overcome the limitations associated with knowledge encoding and generalization in LLMs, making them more adaptable and effective in real-world applications.\\

\subsubsection{\textbf{ZeroShot React Agent}}

Nicholas Crispino, Kyle Montgomery et al. \cite{crispino2023agent} these authors have introduced a new technique that can improve the zero-shot reasoning capabilities of large language models on general language understanding tasks. The technique involves creating an independent agent that guides the reasoning process of the models, thus allowing for greater utilization of their zero-shot reasoning abilities. \\

The performance of a certain method is evaluated by the authors using various datasets that include tasks related to generation, classification, and reasoning. The authors show that their method is highly effective and can be applied to most tasks, as it achieves a state-of-the-art zero-shot performance on 20 out of the 29 evaluated datasets. In particular, their method significantly improves the performance of state-of-the-art LLMs like Vicuna-13b(13.3\%), Llama-2-70b-chat(23.2\%), and GPT-3.5 Turbo(17.0\%), with the average increase in reasoning ability being 10.5\%. \\

The new method's improvement is remarkable compared to the zero-shot chain of thought. For example, using this approach, Llama-2-70b-chat outperforms zero-shot GPT-3.5 Turbo by 10.2\%.\\

\subsubsection{\textbf{Action Executor}}
Action Executor, the primary concept behind using agents involves utilizing a language model to determine a set of actions to execute. In contrast to chains, where actions are predetermined, agents employ a language model as a reasoning engine to decide on the most suitable course of action to take and the sequence in which they should be performed.\\

% -----------------------------------

\begin{itemize}

\item There are a few critical components involved:

\begin{enumerate}[label=\textbf{\roman*.}]

\item \textbf{Schema}\\
There are several crucial elements involved in this Schema:

\begin{enumerate}

\item \textbf{AgentAction}
This statement refers to a particular task that an agent is expected to perform. It has two properties: tool (the name of the tool to invoke) and tool\_input (the input to that tool).
\item \textbf{AgentFinish}
This text describes the output that an agent returns to the user when it has completed its task. The output is stored in a key-value mapping called return\_values, which includes the final agent output. Typically, this output consists of a string that represents the agent's response and is stored under the 'output' key.
\item\textbf{Intermediate Steps}
It is crucial to keep track of the previous agent actions and outputs to avoid duplication of work in future iterations. This is typed as a List[Tuple[AgentAction, Any]]. Note that in practice, observation is often a string, which is why it is left as type Any to be flexible.

\end{enumerate}

\item \textbf{Agent}
This paragraph describes the process of decision-making in an AI-powered assistant. The decision-making process involves a chain that is powered by three components: a language model, a prompt, and an output parser.

Different AI agents have unique ways of prompting, encoding inputs, and parsing outputs. A list of built-in agents is available for reference, and custom agents can be created for greater control.

\begin{enumerate}
    
\item \textbf{Agent Inputs}
An agent's inputs are in the form of a key-value mapping. The mapping must have at least one required key, 'intermediate\_steps', which represents the intermediate steps as mentioned earlier.

The PromptTemplate is responsible for transforming these key-value pairs into a format that is most suitable for inputting into the LLM.

\item \textbf{Agent Outputs}
The output is the next action(s) to take or the final response to send to the user (AgentActions or AgentFinish). Concretely, this can be typed as Union[AgentAction, List[AgentAction], AgentFinish].

The parser transforms raw LLM output into one of three types.

\end{enumerate}

\item \textbf{AgentExecutor}
The agent executor is the crucial component responsible for executing an agent's actions and passing the outputs back to the agent. It plays a significant role in the overall functioning of the agent and ensures that the actions are executed efficiently.

% \begin{lstlisting}[language=Python , title=Pseudocode example]
% next_action = agent.get_action(...)
% while next_action != AgentFinish:
%     observation = run(next_action)
%     next_action = agent.get_action(..., next_action, observation)
% return next_action
% \end{lstlisting}

\SetKwComment{Comment}{/* }{ */}
\begin{algorithm}
\caption{Agent Action Loop}\label{alg:action_loop}
\SetKwInOut{Input}{Input}
\SetKwInOut{Output}{Output}
\Input{Agent, AgentFinish}
\Output{Final Action}
next\_action $\gets$ agent.get\_action(...)\;
\While{next\_action $\neq$ AgentFinish}{
    observation $\gets$ run(next\_action)\;
    next\_action $\gets$ agent.get\_action(..., next\_action, observation)\;
}
\Return{next\_action}\;
\end{algorithm}

The algorithm \ref{alg:action_loop} shows how the Agent works. Although it may seem simple, this runtime handles several complexities for you, including:
\begin{itemize}
    \item[$\blacksquare$] Dealing with cases where the agent chooses a tool that does not exist.
    \item[$\blacksquare$] Handling cases where the tool give errors
    \item[$\blacksquare$] Handling cases where the agent produces output that cannot be parsed into a tool invocation
    \item[$\blacksquare$] Logging and observability at all levels (agent decisions, tool calls) to stdout and/or to LangSmith.
\end{itemize}

\item \textbf{Tools for Collecting and Generating Information}\\
Tools in an agent's system are functions that can be invoked. The Tool abstraction is composed of two components. Firstly, the input schema for the tool specifies what parameters are required to call the tool. This helps the system understand the correct inputs and what they should be named and described as. Secondly, the function to run is generally a Python function that is invoked to perform the desired task.

Two important design considerations should be taken into account when designing tools for an agent's system:
\begin{itemize}
    \item[$\blacksquare$] Giving the agent access to the right tools
    \item[$\blacksquare$] Describing the tools in a way that is most helpful to the agent
\end{itemize}

It is crucial to consider both aspects to create a functional agent. Providing the agent with the appropriate tools is essential to successfully achieve its objectives. Accurate descriptions of these tools are also necessary for the agent to use them effectively. 

LangChain offers a diverse range of pre-built tools and allows users to create their own tools with custom descriptions. To view a comprehensive list of built-in tools, please refer to the tools integrations section.
\begin{enumerate}
    
\item \textbf{Toolkits from LangChain}
LangChain offers a concept of toolkits, which are groups of 3-5 related tools that an agent may require to complete specific tasks. For instance, the GitHub toolkit includes tools for searching through GitHub issues, reading files, commenting, and more. This approach should simplify the process of accomplishing common objectives.

\end{enumerate}

\end{enumerate}
\end{itemize}

% --------------------------------------

\subsubsection{\textbf{LangChain Tools} }
Tools are interfaces that allow an agent to interact with the world. This information is valuable because it can be used to create action-taking systems. The name, description, and JSON schema can prompt the LLM, providing it with the necessary information to specify which action to take. The function call is equivalent to executing the selected action.

It's important to note that the simpler the input to the tool, the easier it is for an LLM to use it. Many agents will only work with tools that have a single string input.

\subsubsection{\textbf{Merged Tools}}

LLM-Enhancer uses merged tools instead of using them individually. The tools that are merged should have a common goal. For eg google search, duckduckgo search , Wikipedia all of these tools are a way to get info from the web. So they could be merged together to get better result. If the tools are used individually then the llm agent executor would select every tools using zero-shot react agent. In this case the context window would be used up easily since every response is fed to the next tool if required. Moreover, each tool may not used at once. \\

\SetKwComment{Comment}{/* }{ */}
\begin{algorithm}
\caption{Merging Multiple Agents Data}\label{alg:merge_agent}
\SetKwInOut{Input}{Input}
\SetKwInOut{Output}{Output}
\Input{Input}
\Output{Merged Search Result}

\

min\_chunks $\gets$ 10\;
chunk\_size $\gets$ 400\;
chunk\_overlap $\gets$ 100\;

\

/*Wrapper function for API*/\;
dd $\gets$ DuckDuckGoSearchRun()\;
wiki $\gets$ WikipediaQueryRun()\;
google $\gets$ GoogleSerperAPIWrapper()\;

\

text\_splitter $\gets$ RecursiveCharacterTextSplitter(chunk\_size, chunk\_overlap)\;

\

/*getting text chunks using text splitter*/\;
texts1 $\gets$ text\_splitter.split\_text(dd)\;
texts2 $\gets$ text\_splitter.split\_text(google)\;
texts3 $\gets$ text\_splitter.split\_text(wiki)\;

\

/*initializing vector database*/\;
collection $\gets$ vector\_chroma\_Database\;

\

/*passing chunks with metadata*/\;
collection $\gets$ tests1 , metadata\;
collection $\gets$ tests2 , metadata\;
collection $\gets$ tests3 , metadata\;

\

res $\gets$ empty\_string()\;

\

/*iterating the top 10 related chunks*/\;
\While{i $\neq$ min\_chunks}{

    res $\gets$ res + collection[i]\;
    i $\gets$ i + 1\;
    
}
   
\Return {res}        
\end{algorithm}

So, in our case, the data from the tools are merged together using Algorithm \ref{alg:merge_agent} code and give a response with multiple data sources data. Then there are Recursive Character Text Splitter which is used to split document data recursively to keep all paragraphs, sentences then words together as long as possible. Then they are made into smaller chunks. These chinks are then passed to vector embedding \cite{chroma}.
These are the tools that we work with in our projects. They are given below:

\begin{itemize}    

\item \textbf{SerperApi for using Google Search:}\\ 
Serper is an affordable API provided by Serper.dev that can be used to retrieve answer box, knowledge graph, and organic results data from Google Search. It consists of two parts: first, the setup, and then the references to the specific Google Serper wrapper.

The Serper API is known for its speed and affordability. It delivers fast search results in just 1-2 seconds, making it an excellent tool to unleash the potential of Google Search. We have integrated this tool into our LLM-ENHANCER, which adds an answer box, knowledge graph, and organic results data from Google Search. This allows us to find the most accurate answer to a user's query.

To merge the Serper API, we have used LangChain as our primary framework. We have imported the Wrapper Utility into our model from LangChain and created a merged search class to search for results from Google.\\

\item \textbf{DuckDuckGo for using DuckDuckGo Search:}\\
LangChain's search functionality includes the use of DuckDuckGo Search, which allows users to search for results on the internet. This tool is integrated into our merged-search class by importing and merging its functionality. The DuckDuckGo search API is utilized to retrieve JSON search results, which are parsed and validated to create a new model based on the input data from keyword arguments. If the input data is invalid, a validation error is raised.\\

\item \textbf{Wikipedia for using the vast information from Wiki:}\\
Wikipedia is an online encyclopedia that is free to use and is written and maintained by a community of volunteers, who are known as Wikipedians. It is available in multiple languages and uses a wiki-based editing system called MediaWiki \cite{wikipedia_integration}. As the most visited reference work in the world, it offers a wealth of information on a wide range of topics. 

We have recently added a Wikipedia Tool from LangChain to our LLM-ENHANCER, which enables users to find relevant information from Wikipedia pages.

\end{itemize}

\subsubsection{\textbf{Calculator}}
This is a straightforward and user-friendly calculator tool created specifically to enable llm to perform a wide range of mathematical calculations with ease and speed. Whether you need to add, subtract, multiply, or divide numbers, this calculator will provide you with accurate results in an instant. 

\subsubsection{\textbf{Splitter}}

When we are working with long documents, it is necessary to divide them into smaller chunks that can fit into your model's context window. However, keeping semantically related pieces of text together can be a challenging task.  LangChain provides various built-in document transformers to help you split, merge, filter, and transform documents.

\subsubsection{\textbf{Embeddings}}
Embeddings are a way for Artificial Intelligence to natively represent a wide variety of data types, making them a perfect fit for use with various AI-powered tools and algorithms. They can represent text, images, and soon audio and video. There are many options available for creating embeddings, whether it be through a locally installed library or by calling an API \cite{chroma_docs}.

Chroma \cite{chroma_docs} provides lightweight wrappers around popular embedding providers, making it simple to use them in your applications. 

In figure \ref{fig:chroma22}, Chroma uses the Sentence Transformers "all-MiniLM-L6-v2" model as its default embedding model. This model can generate document and sentence embeddings that can be utilized for various tasks.

\subsubsection{\textbf{ChromaDB}}

\begin{figure}[h]
  \centering
  \includegraphics[width=1\columnwidth]{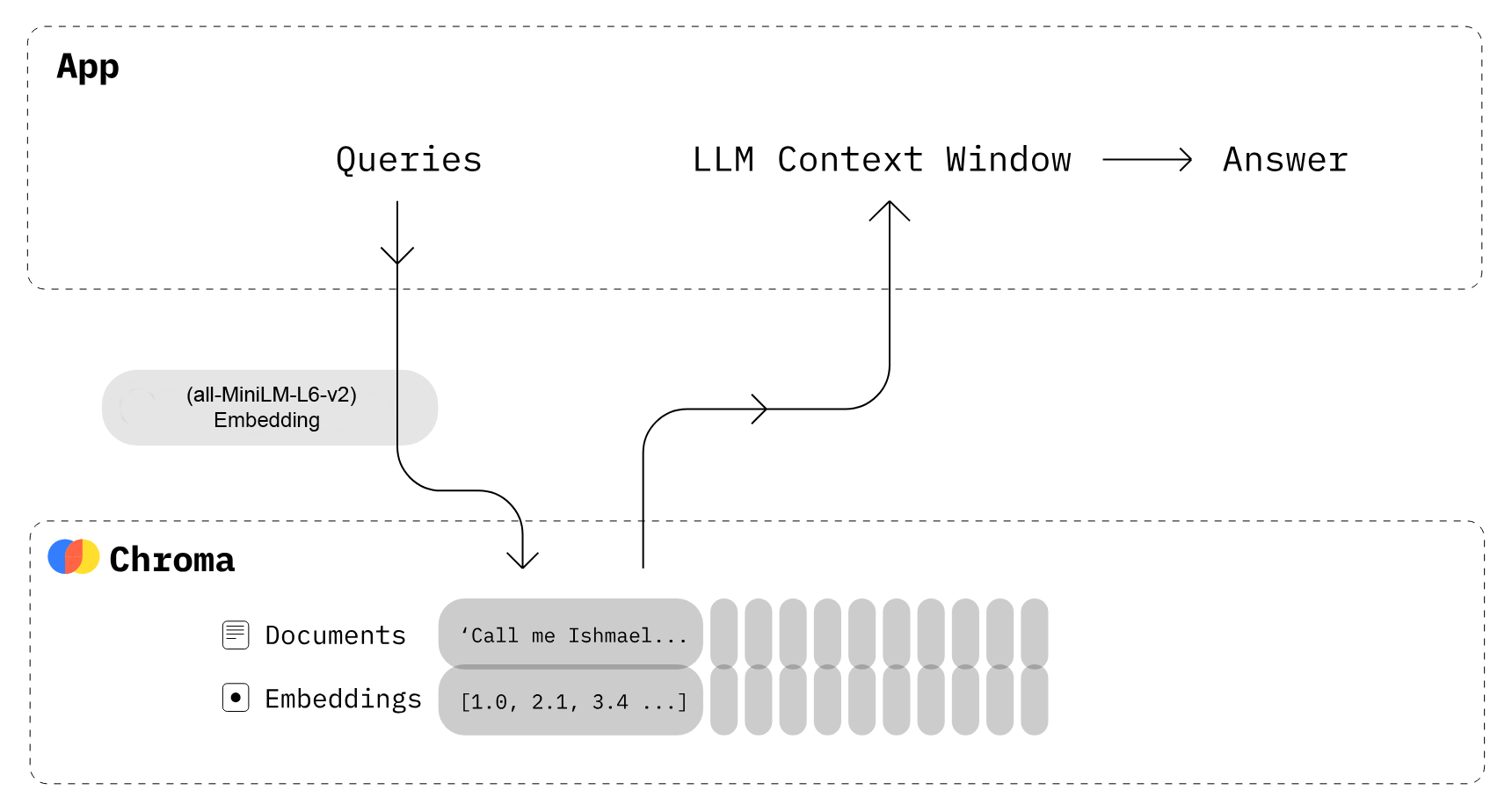}
  \caption{Using Sentence Transformers (all-MiniLM-L6-v2) Embeddings.}
  \label{fig:chroma22}
\end{figure}

In figure \ref{fig:chroma22}, Chroma is an open-source embedding database that simplifies the process of building LLM apps by providing a plug-and-play system for knowledge, facts, and skills. With Chroma's user-friendly platform, you can easily store embeddings and their metadata, embed documents and queries, and search through embeddings.

We use vector embeddings to store and easily search relevant chunks of data received from search tools.

\subsubsection{\textbf{Mistral 7B (Opensource LLM)}}

Mistral 7B is the latest 7B model that offers better performance and efficiency. It incorporates Grouped-query attention (GQA) and Sliding Window Attention (SWA) mechanisms, these allow for quicker processing and management of lengthier sequences while reducing costs.\\
Mistral 7B surpasses Llama 2 13B in all benchmarks and even comes close to Llama 1 34B in many benchmarks. It performs exceptionally well on code-related tasks while maintaining proficiency in English tasks.\\
It is available under the Apache 2.0 license, which allows unrestricted use. You can download and use it locally with their reference implementation, deploy it on different cloud platforms using the vLLM inference server and sky pilot, or use it on HuggingFace.

\vspace{1cm}
\section{System Implementation}

In this section of our paper, we will discuss our Answer Comparator(for evaluating our LLM-ENHANCER) and the datasets we are working with.\\

\subsection{Answer Comparator}
Answer Comparator, which is an automated system pipeline for comparing answers. It is also powered by LLM. When an answer is generated from LLM-ENHANCER using a custom prompt, it is evaluated. The custom prompt template is given in Code 1 to check and evaluate the answer of LLM with real answers.

\begin{lstlisting}[title=Code 1. Prompt template for Comparison]
template_for_comapring_answers = """

Read the following Prompt1 and Prompt2 and determine if Prompt2 is available in Prompt1



Prompt1: "{llm_generated}"

Prompt2: "{actual_answer}"

Options:

Yes, the answer is available in the Prompt1 .
No, the answer is not available in the Prompt2 sentence.

"""
\end{lstlisting}

\begin{figure}[h]
  \centering
  \includegraphics[width=0.65\columnwidth]{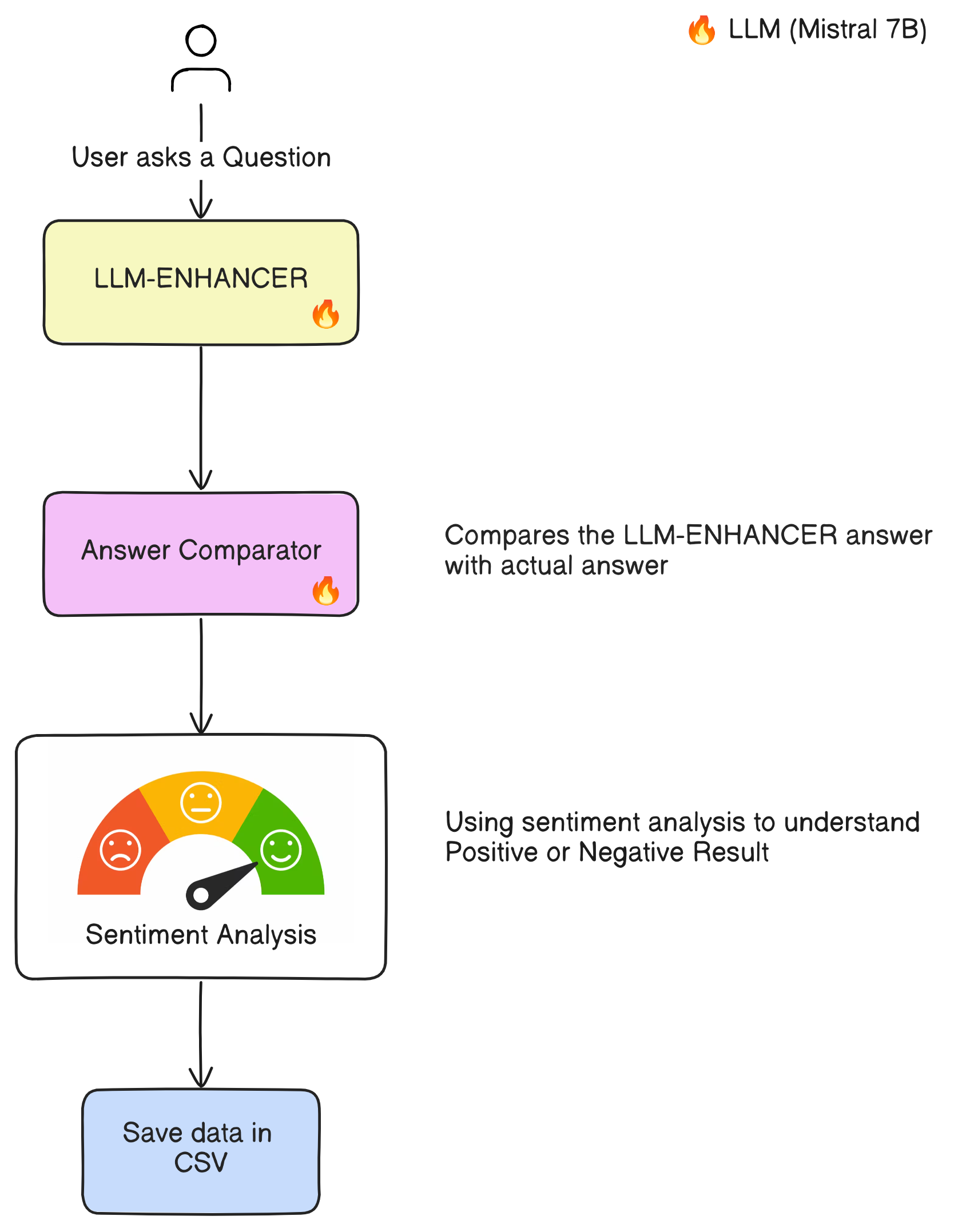}
  \caption{Answer Comparator.}
  \label{fig:Answer Comparator}
\end{figure}

From Code 1 the output is passed to the sentiment analysis model to get the sentiment value. Observing the sentiment value, we can be sure the results match the actual result. Then, the data is saved in the CSV file.
Here the "distilbert-base-multilingual-cased-sentiments-student" model is being used for sentiment analysis.\\
In Figure \ref{fig:Answer Comparator}, we have provided the model of our Answer Comparator.

Figure \ref{fig:Answer Comparator} is the architecture for an automated pipeline for comparing correct answers. It uses Symentic Analysis for Similarity checking the Answers of LLM and real answers.

\subsection{Dataset}

We have used "WikiQA" \cite{yang-etal-2015-wikiqa} Dataset in our system to evaluate. We also have a Recent dataset consisting of recent Question and Answer from 2023-24 recent events that happened. We need to preprocess the data to extract the question-answer column.

\subsubsection{Preprocessing}
\begin{figure}[h]
  \centering
  \includegraphics[width=0.8\columnwidth]{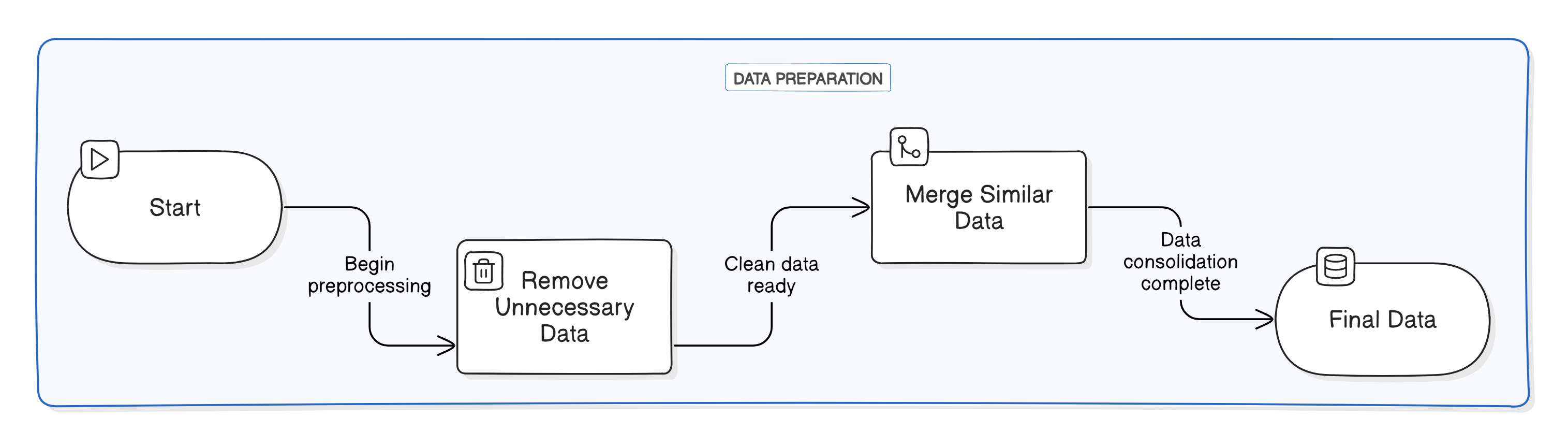}
  \caption{Data Preprocessing}
  \label{fig:Preprocessing}
\end{figure}

We have removed unnecessary data and combined all the required data to create a modified dataset that can be fed into our model Figure \ref{fig:Preprocessing}. It is not possible to use the WikiQA dataset in our proposed technique without making specific alterations to it. Therefore, we have modified the data as illustrated in the tables below.

\vskip2ex

\subsubsection{WikiQA Dataset (preprocessed)}

The WikiQA corpus is a publicly available collection of questions and sentence
pairs that have been marked for research purposes related to open-domain question answering.
We have made modifications and preprocessing to this dataset to test our
model. We have selected specific fields to assess our model's performance:
\begin{itemize}
    \item question
    \item answer
\end{itemize}
Here is the demo question and answer from this dataset :
\vskip1ex
\begin{figure}[h]
  \centering
  \includegraphics[width=0.8\columnwidth]{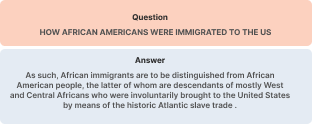}
  \caption{Demo Question of WikiQA Dataset (preprocessed)}
  \label{fig:demo55}
\end{figure}
\vskip2ex
From Figure \ref{fig:demo55} and Table  \ref{tab:glacier_caves1}, we can see that the dataset Mostly consists of questions that ask for definitions or general knowledge. No recent data is available for the years 2022, 2023, and 2024 in this dataset. Most of the data here was gathered up until. Using this dataset to evaluate our model is not wise because our main goal is to enhance the external knowledge of our LLMs.
We compared the performance of our Sequential and Merge models on definition-type questions using this dataset.
\\
After preprocessing the WikiQA dataset, we have 369 questions and their corresponding answers. Figure \ref{fig:Preprocessing} illustrates the steps necessary for data pre-processing.

\begin{table}[htbp]
\centering
\caption{Structure of WikiQA train dataset}
\begin{tabular}{|c|p{1cm}|c|p{1cm}|p{1cm}|}
\hline
\textbf{question\_id} & \textbf{question} & \textbf{document\_title} & \textbf{answer} & \textbf{label} \\ \hline
Q1 & how are glacier caves formed? & Glacier cave & A partly submerged glacier cave on Perito Moreno Glacier. & 0 \\ \hline
Q1 & how are glacier caves formed? & Glacier cave & The ice facade is approximately 60 m high. & 0 \\ \hline
Q1 & how are glacier caves formed? & Glacier cave & Ice formations in the Titlis glacier cave. & 0 \\ \hline
Q1 & how are glacier caves formed? & Glacier cave & A glacier cave is a cave formed within the ice of a glacier. & 1 \\ \hline
Q1 & how are glacier caves formed? & Glacier cave & Glacier caves are often called ice caves, but this term is properly used to describe bedrock caves that contain year-round ice. & 0 \\ \hline
\end{tabular}

\label{tab:glacier_caves1}
\end{table}

In Table \ref{tab:glacier_caves1} we can see the how the datas are represented.
To evaluate our LLM, we only need the questions labeled as 1. If there are multiple questions with the same ID, we can merge them.

\begin{table}[htbp]
\centering
\caption{Structure of preprocessed WikiQA train dataset}
\begin{tabular}{|c|p{4cm}|}
\hline
\textbf{Question} & \textbf{Answer} \\ \hline

how are glacier caves formed? & A glacier cave is a cave formed within the ice of a glacier.  \\ \hline

\end{tabular}

\label{tab:glacier_caves2}
\end{table}

Table \ref{tab:glacier_caves2} shows the final table that we will use.
Now the unique value in the dataset is reduced.

\begin{table}[htbp]
\centering
\caption{Unique values in WikiQA dataset}
\begin{tabular}{|c|p{3cm}|p{3cm}|}
\hline
\textbf{} & \textbf{Before Preprocessing} & \textbf{After Preprocessing}\\ \hline

train & 20360 & 873 \\ \hline
test & 6165 & 243 \\ \hline
validation & 2733 & 126 \\ \hline

\end{tabular}

\label{tab:glacier_caves3}
\end{table}

Since we don't have to train we can use 243+126=369 unique values for evaluation in Table  \ref{tab:glacier_caves3}.

As we know, most of the LLMs have been trained with data up until 2021, and the wikiQA dataset is from 2015 so this has been trained for the LLM. Which is not ideal for evaluating the performance of LLM.
So we have also made a dataset of our own which consists of 500 unique data from 2023-2024 available on the web.

\subsection{Questions Dataset2023-24}
We have also created another dataset containing 500 unique values and data from 2023-2024.
In our dataset, we have questions about recent activities, calamities, sports 
news, news, social activities, Wikipedia-based information, TV shows, and 
general knowledge until 2024.
We have used the information from these sources to take answers which will be 
used to evaluate our LLM model's response.\\
\\
\begin{itemize}
    \item Google
    \item Wikipedia
    \item DuckDuckGO
\end{itemize}

\begin{table}[htbp]
\centering
\caption{Structure of Dataset2023-24 dataset}
\begin{tabular}{|p{4cm}|l|}
\hline
\textbf{Question} & \textbf{Answer} \\ \hline

who is Ryan Michael Blaney's grandfather?	&Lou Blaney \\ \hline
Ryan Michael Blaney started his racing career with what type of racing?	& Quarter midget racing \\ \hline
who got Nobel Prize 2023 in Peace ?	& Narges Mohammadi \\ \hline
who won 2023 Cricket World Cup?	&Australia\\ \hline
Who is the player of the series in 2023 Cricket World Cup?	&Virat Kohli\\ \hline
How many teams are their in 2023 Cricket World Cup?	&10 teams\\ \hline
Bangladesh won how many matches in 2023 Cricket World Cup Mens ODI?	&2 matches\\ \hline
Which Bangladeshi teenager nominated for International Children's Peace Prize 2023?	&Mainul Islam\\ \hline
Who was the interim CEO to lead OpenAI in 2023 when Sam Altman was fierd?	&Mira Murati\\ \hline

\end{tabular}

\label{tab:glacier_caves}
\end{table}

\begin{figure}[h]
  \centering
  \includegraphics[width=0.8\columnwidth]{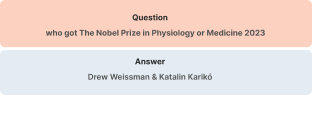}
  \caption{Demo Question of Dataset2024}
  \label{fig:ex122}
\end{figure}

Our Dataset2024 comprises 500 recently collected questions and answers from these various sources like represented in Table \ref{tab:glacier_caves}.
Data Fields are in our Dataset2024:
\begin{itemize}
    \item question
    \item answer
\end{itemize}
We can compare our Offline LLM and Online LLM with Sequential and Merged Tools
with this dataset because it contains recent data up to 2024.
Here is the demo question and answer from this dataset in Figure \ref{fig:ex122}:
\vskip1ex
\vskip2ex

\subsection{System Requirements}
 It presents an overview of different approaches for identifying user requirements and system specifications and concludes with a short description. System software and hardware requirements to work with our models. These are our specifications given Below:

\begin{itemize}
    \item Processor: Intel i5 13th gen 14 core
    \item RAM: 32 GB 3200MHZ
    \item Storage: 512 GB NVME SSD
    \item Operating System: Windows 10
    \item Graphics Card: NVIDIA RTX 3060 12 GB
\end{itemize}

% \section{Evaluation Metrics}

%=========================================================

\vspace{1cm}
\section{Performance Evaluation}

In this section of the paper, we will discuss the outcomes of comparing the LLM with the sequential tool model versus LLM with the merge tool(LLM-ENHANCER) approach where vector embedding is used.

\subsection{Large Language Models Used}
\begin{figure}[h]
  \centering
  \includegraphics[width=0.90\columnwidth]{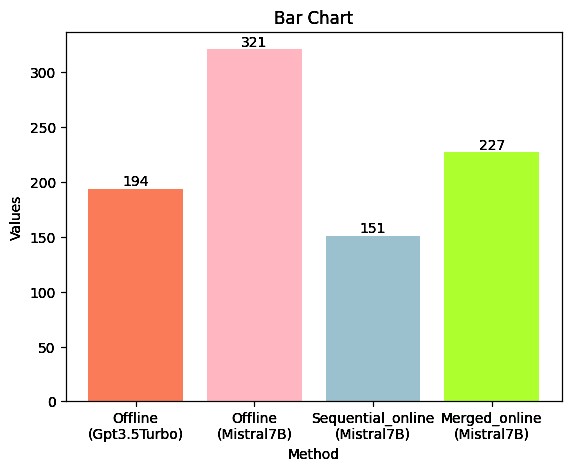}
  \caption{Amount of right answer using WikiQA dataset using Different approaches}
  \label{fig:5.1}
\end{figure}

In Figure \ref{fig:5.1}, we evaluated 4 types of approaches using the old dataset WikiQA \cite{yang-etal-2015-wikiqa} 2015. These models are:
\begin{itemize}
    \item LLM(Gpt3.5Turbo)
    \item LLM(Mistral7B)
    \item LLM(Mistral7B) + Sequential Online Tools
    \item LLM(Mistral7B) + Merged Online Tools(LLM-ENHANCER)
\end{itemize}
Please note that the dataset named wikiQA, which we're using, has questions only up until 2015. Although this dataset is quite common and is used to train various LLM models, it can be used to benchmark the proposed approach.

\vskip2ex

\vskip1ex
\begin{figure}[h]
  \centering
  \includegraphics[width=0.90\columnwidth]{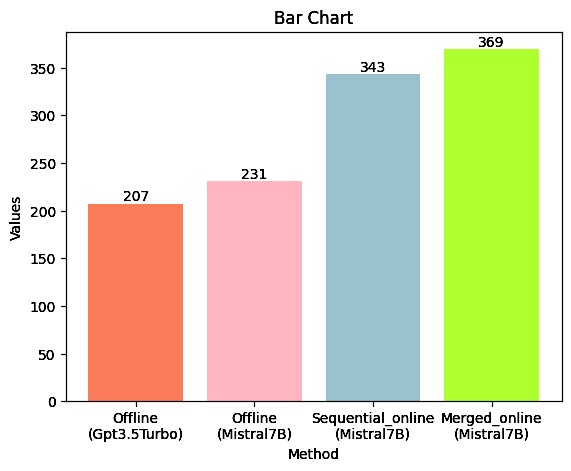}
  \caption{Amount of right answer using recent Dataset2023-2024 dataset using Different approaches}
  \label{fig:5.3}
\end{figure}
\vskip2ex

 In figure \ref{fig:5.1}, we can see that Offline(Gpt3.5Turbo) and Offline(Mistral7B) works much better than Sequential\_online(Mistral7B) which is connected to the Internet. The reason behind it is that since these models were trained on large corpus of data and WikiQA data set is the most common dating from 2015. The LLMs already know the answer regarding it. But Our Merged Approach is able to beat Offline(Gpt3.5Turbo). Our technique is able to solve 227 questions.\\
 
 We can see that the Offline approach works better than others. However, the key point is that our merged LLM-ENHANCER approach works significantly better.\\

In Figure \ref{fig:5.3}, we can see that we have used our Dataset2023-24, which consists of 500 unique values. In this case, our merged LLM-ENHANCER approach works significantly better. As shown we have 5.2\% improvement from the sequential approach and 32.4\% improvement from GPT3.5Turbo.
\\
\\
\\
\vskip1ex
\begin{figure}[h]
  \centering
  \includegraphics[width=0.90\columnwidth]{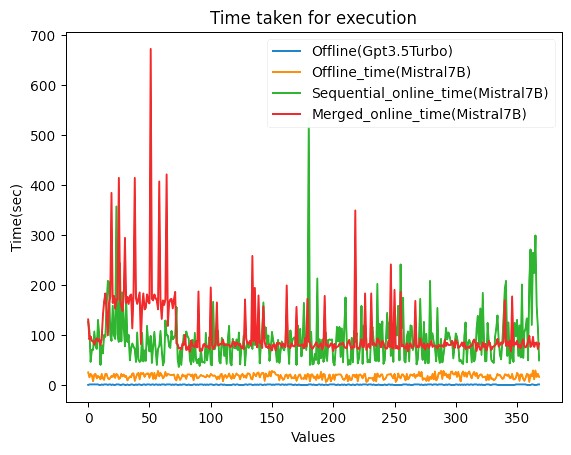}
  \caption{Time to execute each question in WikiQA dataset using Different approaches}
  \label{fig:5.2}
\end{figure}
\vskip2ex

\vskip1ex
\begin{figure}[h]
  \centering
  \includegraphics[width=0.90\columnwidth]{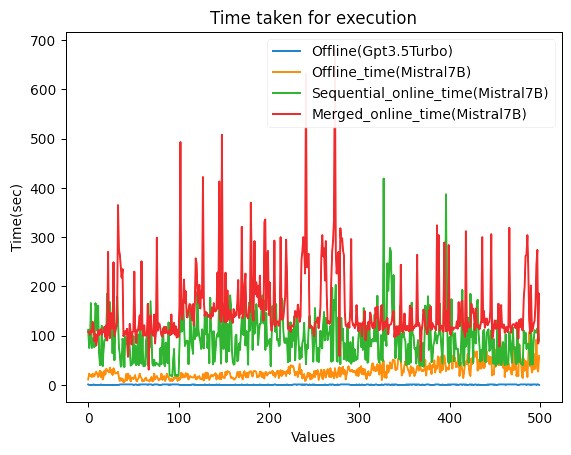}
  \caption{Time to execute each question in Dataset2024 using Different approaches }
  \label{fig:5.4}
\end{figure}
\vskip2ex

In Figure \ref{fig:5.2}, we can see that the offline approach takes less time to execute than LLM with the sequential tools and our LLM enhancer with merge tools. Offline LLM gives better results also.\\
Offline LLM is trained with the old dataset until 2021, so it can easily solve older datasets like wikiQA even when offline. However, when we provide it with a new recent dataset, its performance worsens, as shown in Figure \ref{fig:5.4}.

For classification problems, accuracy is often misleading if the classes are imbalanced. Several other measures, such as precision, recall, F1-score, training time, testing time, and confusion matrix, are used to evaluate the performance of our proposed model. The following defines these performance matrices:

\subsubsection{Precision}
Precision measures the proportion of accurately predicted positive values to the total number of predicted positives.
\begin{align}
Precision = \frac{TP}{TP+FP}  
\end{align}

\subsubsection{Recall}
The sensitivity or detection rate, known as recall, measures the proportion of actual positive values that are correctly predicted.
\begin{align}
Recall = \frac{TP}{TP + FN}   
\end{align}

\subsubsection{F1-score}
The F1-score is a metric that calculates the harmonic mean of precision and recall for classification problems. It provides a balanced measure of both precision and recall, where precision measures the accuracy of positive predictions and recall measures the coverage of actual positive instances.
\begin{align}
F1-score = \frac{2(Precision * Recall)}{Precision + Recall}   
\end{align}
\\
LLM-ENHANCER takes longer to run as vector embeddings need to be built each time. But the accuracy is worth it.
\\

\begin{table}[h]
\centering
\caption{WikiQA Dataset 369 unique Data}
\small % Reduce font size
\begin{tabular}{|l|p{1.2cm}|p{0.9cm}|p{1.4cm}|p{1.5cm}|}
\hline
\textbf{Metrics}& \textbf{GPT 3.5 Turbo} & \textbf{Mistral 7b} & \textbf{Sequential Model} & \textbf{LLM Enhencer}\\
\hline
\hline
Precision & 0.52 & \textbf{0.87} & 0.41 & \textbf{0.62}\\
\hline
Recall & 1 & 1 & 1 & 1\\
\hline
F1 Score & 0.68 & \textbf{0.93} & 0.58 & \textbf{0.77}\\
\hline
\end{tabular}

\label{tab:5.1}
\end{table}

We can see from table \ref{tab:5.1} that the F1 score of mistral7b is higher since it was trained on those data. But we have F1 score of 0.77 from the 0.58 sequential technique which is a 19\% improvement. It is worth noting that the paper \cite{peng2023check} used GPT3 and got precision score a little bit lower which is 0.48 rather than 0.52. So the test technique is appropriate in this regard.

\begin{figure}[h]
  \centering
  \includegraphics[width=0.90\columnwidth]{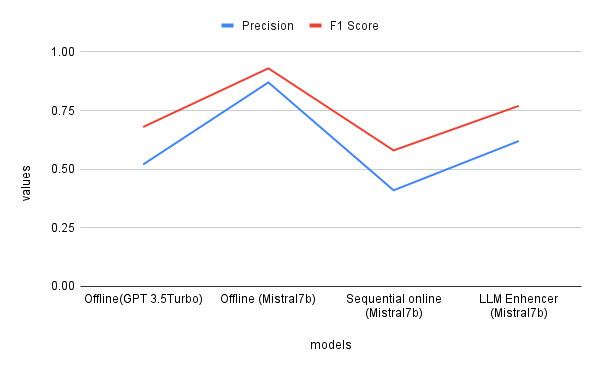}
  \caption{WikiQA dataset precision and recall comparison}
  \label{fig:16}
\end{figure}
\vskip2ex

We can see the values of table \ref{tab:5.1} on a line graph presented in Fig. \ref{fig:16}. Here we only see an anomaly with the mistral model.\\

\begin{table}[h]
\centering
\caption{ Dataset2024 500 unique recent Data}
\small % Reduce font size
\begin{tabular}{|l|p{1.2cm}|p{0.9cm}|p{1.4cm}|p{1.5cm}|}
\hline
\textbf{Metrics}& \textbf{GPT 3.5 Turbo} & \textbf{Mistral 7b} & \textbf{Sequential Model} & \textbf{LLM Enhencer}\\
\hline
\hline
Precision & 0.41 & 0.46 & 0.69 & \textbf{0.74}\\
\hline
Recall & 1 & 1 & 1 & 1\\
\hline
F1 Score & 0.58 & 0.63 & 0.82 & \textbf{0.85}\\
\hline
\end{tabular}

\label{tab:5.2}
\end{table}

In Table \ref{tab:5.2}, the dataset is entirely new to the LLM model.\\
\\
\\

\begin{figure}[h]
  \centering
  \includegraphics[width=0.90\columnwidth]{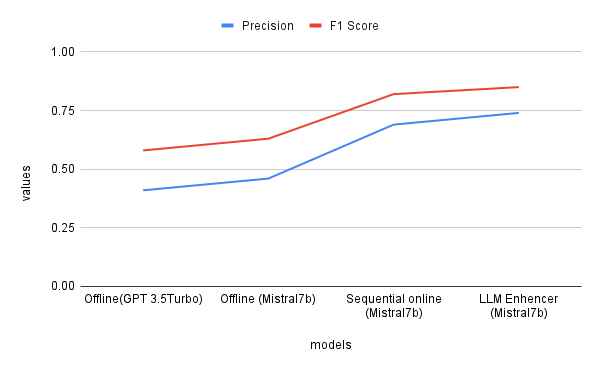}
  \caption{Dataset2024 dataset precision and recall comparison}
  \label{fig:17}
\end{figure}
\vskip2ex

We can see the values of table \ref{tab:5.1} on a line graph presented in Fig. \ref{fig:16}.  According to Table \ref{tab:5.2}, the LLM-ENHANCER F1 score has increased by 22\% to 0.84 from 0.63.
We can also see 27\% increase from GPT3.5Turbo.\\

This technique has shown great potential in our testing . This will help in enhancing LLMs largely.

%=========================================================

\vspace{1cm}
\section{Conclusions and Future Work}

\subsection{Summary of Research}

Large language models (LLMs) like ChatGPT exhibit human-like abilities in generating natural responses across various tasks such as task-oriented dialog and question answering. However, deploying LLMs in critical real-world applications poses challenges due to their tendency to generate false information, lack of external knowledge utilization, and reliance on outdated data. To address these challenges, this paper introduces the LLM-ENHANCER system, which leverages multiple online sources like Google, Bing, Wikipedia, and DuckDuckGo to acquire more accurate data.

The LLM-ENHANCER system employs open-source LLMs and custom tools for parallel data ingestion, rather than serial processing. By utilizing vector embeddings, the system identifies the most relevant information from the amalgamated data sources and feeds it to the LLM for generating responses to user queries. The primary aim of LLM-ENHANCER is to mitigate the occurrence of chat-LLM hallucinations, where the model generates erroneous or misleading information while ensuring that the responses remain both natural and accurate.

Through experimental evaluation, LLM-ENHANCER demonstrates a significant reduction in chat-LLM hallucinations while maintaining the quality and authenticity of responses. Furthermore, the paper emphasizes the accessibility of the source code and models, which are made publicly available, encouraging transparency and further development in the field of large language models.

\subsection{Limitations}
One of the main limitations of using the merged approach is the token size. For instance, in the sequential technique, the agent calls the Google tool, which returns the answer in 20 tokens. However, in the merged approach, it merges data from multiple sources, such as Google (20 tokens), DuckDuckGo (50 tokens), and Wikipedia (100 tokens), resulting in a total of 170 tokens. This means that the merged approach will take significantly more time to process the data than the sequential technique. Moreover, a better embedding model could be used for much faster data retrieval. One of the bottlenecks of this system is vector embedding itself. Embedding data to vector takes time, slowing the entire operation and hampers the speed. However, it will be more accurate.

\subsection{Future Work}
In the future, we can achieve better accuracy by applying more advanced vector embedding models and language models. The chunk size and overlapping can be configured to check performance metrics.

% \section*{References}

\end{document}